\documentclass[sn-mathphys]{sn-jnl}

\usepackage{booktabs}
\usepackage{hyperref}
\usepackage[utf8]{inputenc}
\usepackage{multirow}
\usepackage{subcaption}
\usepackage{rotating}
\usepackage[normalem]{ulem}
\usepackage{xcolor}

\newcommand{\doubleplus}{\ensuremath{+\!\!\!\!+\,}}

\jyear{2023}

\begin{document}

\title[Traffic Forecasting on New Roads Using SCPT]{Traffic Forecasting on New Roads Using Spatial Contrastive Pre-Training (SCPT)}


\author[1,2]{\fnm{Arian} \sur{Prabowo}}\email{arian.prabowo@gmail.com}
\author[3]{\fnm{Hao} \sur{Xue}}\email{hao.xue1@unsw.edu.au}
\author[2]{\fnm{Wei} \sur{Shao}}\email{wei.shao@data61.csiro.au}
\author[2,4]{\fnm{Piotr} \sur{Koniusz}}\email{piotr.koniusz@data61.csiro.au}
\author*[3]{\fnm{Flora} \sur{Salim}}\email{flora.salim@unsw.edu.au}

\affil*[1]{\orgdiv{Computing Technologies}, \orgname{RMIT}, \orgaddress{\city{Melbourne}, \postcode{3000}, \state{VIC}, \country{Australia}}}

\affil[2]{\orgdiv{Data61}, \orgname{CSIRO}, \orgaddress{ \city{Canberra}, \postcode{2601}, \state{ACT}, \country{Australia}}}

\affil[3]{\orgdiv{Computer Science and Engineering}, \orgname{UNSW}, \orgaddress{ \city{Sydney}, \postcode{2052}, \state{NSW}, \country{Australia}}}

\affil[4]{\orgdiv{Engineering, Computing and Cybernetics}, \orgname{ANU}, \orgaddress{ \city{Canberra}, \postcode{2600}, \state{ACT}, \country{Australia}}}

\abstract{
New roads are being constructed all the time.
However, the capabilities of previous deep forecasting models to generalize to new roads not seen in the training data (unseen roads) are rarely explored.
In this paper, we introduce a novel setup called a spatio-temporal (ST) split to evaluate the models' capabilities to generalize to unseen roads.
In this setup, the models are trained on data from a sample of roads, but tested on roads not seen in the training data.
Moreover, we also present a novel framework called Spatial Contrastive Pre-Training (SCPT) where we introduce a spatial encoder module to extract latent features from unseen roads during inference time.
This spatial encoder is pre-trained using contrastive learning.
During inference, the spatial encoder only requires two days of traffic data on the new roads and does not require any re-training.
We also show that the output from the spatial encoder can be used effectively to infer latent node embeddings on unseen roads during inference time.
The SCPT framework also incorporates a new layer, named the spatially gated addition (SGA) layer, to effectively combine the latent features from the output of the spatial encoder to existing backbones.
Additionally, since there is limited data on the unseen roads, we argue that it is better to decouple traffic signals to trivial-to-capture periodic signals and difficult-to-capture Markovian signals, and for the spatial encoder to only learn the Markovian signals.
Finally, we empirically evaluated SCPT using the ST split setup on four real-world datasets.
The results showed that adding SCPT to a backbone consistently improves forecasting performance on unseen roads.
More importantly, the improvements are greater when forecasting further into the future.
The codes are available on GitHub: \burl{https://github.com/cruiseresearchgroup/forecasting-on-new-roads}.
}


\maketitle

\section{Introduction}

\begin{figure}
    \centering
    \includegraphics[width=\textwidth]{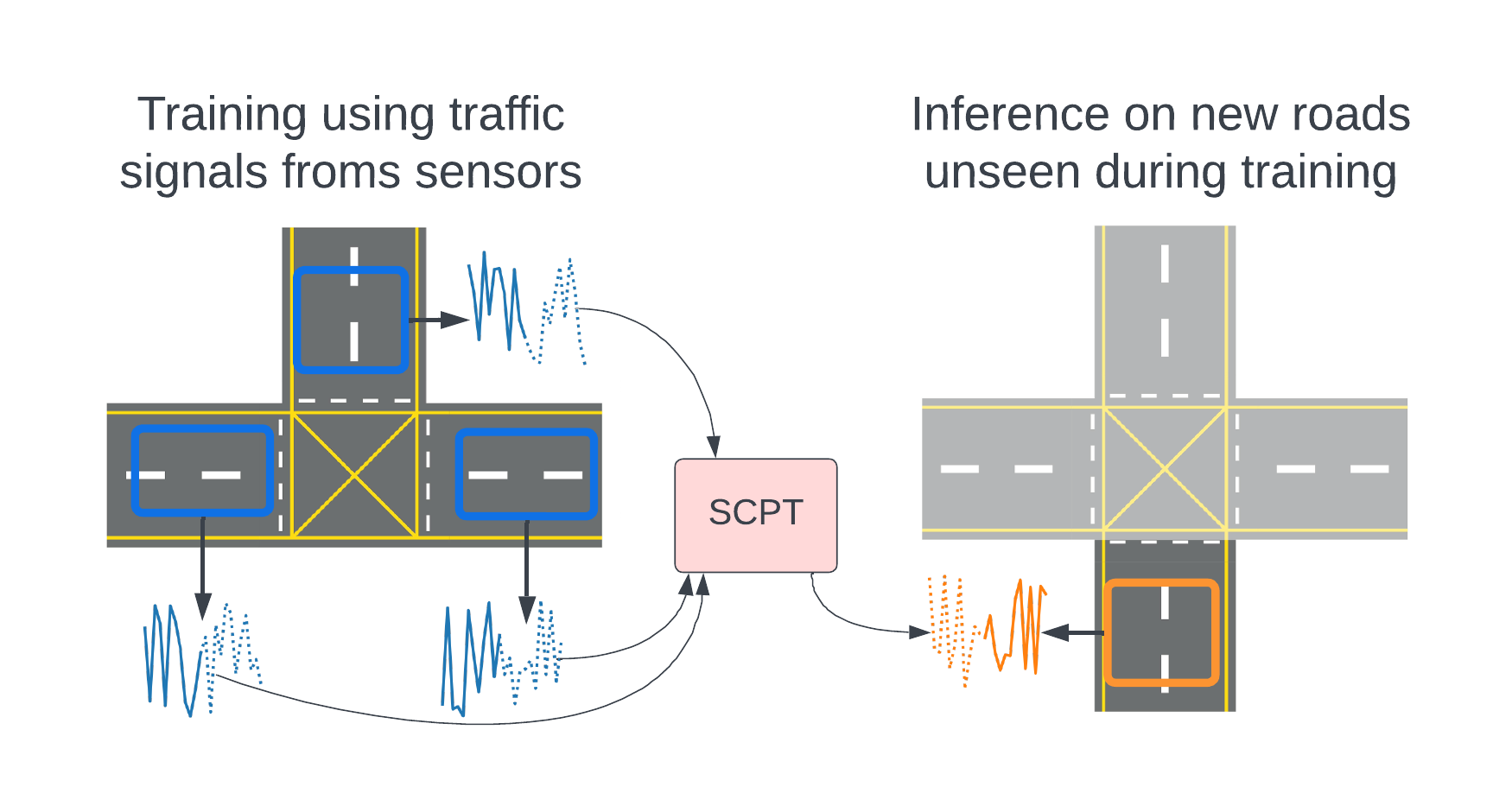}
    \caption{
    Our novel traffic forecasting framework, Spatial Contrastive Pre-Training (SCPT), enables accurate forecasts on new roads (orange) that were not seen during training.
    }
    \label{fig:vizabs}
\end{figure}

Traffic forecasting is a critical component of intelligent transport systems.
It enables the proactive management of traffic congestion and the efficient utilization of limited resources such as road space and public transportation.
Road networks also changes through time as newly constructed roads are being added to the existing networks.
Despite its importance, traffic forecasting on these new roads are not well explored \cite{roth2022FUNn}.

The naive approach is to re-train the models with new data whenever new roads are constructed.
This is not ideal for a two reasons.
Firstly, re-training for an entire network is expensive.
For example, training a model on one third of the entire California highway network took nearly 8 GPU-days \cite{mallick2020gpdcrnn}.
Secondly, new roads, by their nature, have limited data, complicating the re-training process.
Therefore, there is a need for methods to evaluate and extend the capabilities of a trained model to generalize to new roads not seen in the training data (unseen roads).

For evaluation, we introduced a novel setup to split data into train-validate-test sets to evaluate a model's capability to generalize to unseen roads.
We refer to this setup as a spatio-temporal (ST) split because it incorporates both the temporal and spatial aspects of the data. In addition to reserving future time for validation and testing, we also allocate a portion of the roads (representing the spatial aspect) for evaluation. This allows us to assess the model's performance in terms of both time-based forecasting accuracy and its ability to generalize to new spatial locations within the network.
This setup is data-driven, meaning it does not require contextual information about these new roads, e.g., road types, speed limits.
Instead, ST splits only requires minimal traffic data on the new roads and does not require any re-training.
The traffic data required is very short, two days as opposed to few months that is typical to the traffic forecasting tasks \cite{jiang2021dltraff}.


For methods aimed at extending the capabilities of trained models to generalize to new, unseen scenarios, unsupervised pretraining such as contrastive learning has emerged as a promising approach.
Unsupervised pretraining involves training an encoder on a large amount of unlabeled data to learn useful representations without relying on explicit labels. One popular technique within unsupervised pretraining is contrastive learning, which encourages the encoder to capture meaningful features by contrasting positive and negative samples.

In domains such as natural language processing \cite{brown2020GPT3}, audio \cite{oord2018CPC}, and images \cite{chen2020imageGPT} \cite{chen2020simclr} the adoption of unsupervised pretraining has led to the development of powerful encoders capable of achieving generalized performance on diverse downstream tasks.
Surprisingly, the application of unsupervised frameworks to traffic forecasting remains limited \cite{liu2022when} \cite{li2022SPGCL} with zero exploration of leveraging contrastive learning techniques to enhance model generalization to previously unseen roads.

In this paper we proposed a novel framework called \textbf{Spatial Contrastive Pre-Training (SCPT)} illustrated in Figure \ref{fig:vizabs}.
Here, we added a new pre-training stage where the spatial encoder is exposed to the entire historical traffic signal of a road in the training set and then tasked to minimize the contrastive loss between latent representation of different roads (blue boxes).
The models are only trained on the data from the roads in the training set (blue timeseries).
They also take as input the output of the spatial encoders.
During inference, forecasting is performed on traffic signals (orange timeseries) from new sensors (orange box) on roads which are not previously seen during training.

Besides the contrastive pre-training, we also introduce three techniques in the SCPT framework to increase its effectiveness.
The first technique is a spatially gated addition (SGA) layer that uses a gating mechanism to integrate the output of the spatial encoders with the backbone forecasting models more effectively.
Furthermore, it is important to consider the decomposition of traffic signals into two distinct components: trivial-to-capture periodic signals and difficult-to-capture Markovian signals. If only the raw traffic signal is used for contrastive learning, the model may primarily rely on the easily captured periodic signals, while disregarding the more challenging Markovian signals. This is why we have decoupled these two types of signals and employed a separate method to model the periodic signals, ensuring that the model effectively learns the intricate Markovian signals.
Finally, many forecasting models use latent node embeddings that are learned during training \cite{wu2019GraphWaveNet} \cite{DBLP:conf/nips/0001YL0020AGCRN}, making it is impossible to generalize these node embeddings on unseen roads during inference time.
We show that the output of the spatial encoders can be used to effectively infer the node embeddings on unseen roads during inference time.

To empirically evaluate our proposed framework, we use the current state-of-the-art model, called Graph WaveNet (GWN) \cite{wu2019GraphWaveNet} as our forecasting model backbone.
We then implemented the SCPT framework, using the ST split setup, on four real-world traffic datasets, including PeMS-11k, the largest publicly available dataset used for deep traffic forecasting.
The results showed that the SCPT framework consistently improved the efficiency of the backbone.

The main contributions of this paper are:

\begin{itemize}
    \item A new data splitting strategy called a \textbf{ST split}. This allows the evaluation of a framework's capability to perform traffic forecasting on unseen roads.
    \item A novel framework called \textbf{SCPT} that uses contrastive pre-training to allow forecasting models to generalize to unseen roads during inference time.
    \item Empirical evidences from extensive experiments on four real world datasets to gain insights to the effectiveness of SCPT and it's components.
\end{itemize}

\section{Related works}

\subsection{Statistical and machine learning approaches}

The earliest work in traffic forecasting can be classified as data-driven and statistical approaches to machine learning, starting from the Box-Jenkins technique \cite{ahmed1979analysis}. Others included the autoregressive integrated moving average (ARIMA) \cite{hamed1995short} and ARIMA-like approaches such as KARIMA \cite{van1996combining}, subset ARIMA \cite{lee1999application}, ARIMAX \cite{williams2001multivariate}, VARMA \cite{kamarianakis2003forecasting}, and SARIMA \cite{williams2003modeling}. There are also classical machine learning methods such as the support vector regression \cite{jeong2013supervised,lippi2013short,chen2012retrieval}.

\subsection{Early deep learning models}

In the context of traffic forecasting, the advances in deep learning mainly focused on coming up with more sophisticated architectures to improve only the models' accuracy, neglecting other research goals such as actionability and explainability \cite{manibardo2021doesitmakeadifference}.

The early architectures used to better capture the temporal dynamics included stacked autoencoders \cite{lv2014traffic}, gated recurrent units (GRU) \cite{fu2016using}, and long short-term memory (LSTM) \cite{shao2020deep}\cite{cui2020graph}.

Starting from STGCN \cite{yu2018STGCN} and DCRNN \cite{li2018dcrnn_traffic}, the general architecture of choice to capture the traffic spatiotemporal dynamic is to alternate various spatial and temporal modules.
STGCN \cite{yu2018STGCN} alternated between convolutional neural networks (CNN) and spectral graph convolution networks (GCN)\cite{defferrard2016ChebNet}.
DCRNN \cite{li2018dcrnn_traffic} used GRU and diffusion convolution.

Notice that similar the spatiotemporal analysis has also been applied to problems such as prediction of road networks \cite{PrabowoArian2019CCTN}, flight delay prediction \cite{flight_delay,shao2022predicting}, and energy use forecasting \cite{prabowo2023continually} .


\subsection{Current deep learning models}

In the subsequent discussion, we will delve into the latest progress in deep learning models for traffic forecasting and their applicability to our present study.

GWN \cite{wu2019GraphWaveNet} used a CNN called WaveNet \cite{van2016wavenet} and also diffusion convolution from DCRNN.
Moreover, GWN argued that there are hidden spatial dependencies that are not captured by adjacency matrices constructed from physical road networks.
Instead, it argued that the latent topological connectivity should be learned from data
This remained an open challenge until today \cite{wu2020MTGNN, shang2021GTS, li2022SPGCL}.
Our proposed framework, SCPT, extended this capability by being able to infer the latent topological connectivity of new roads which are unseen in the training data during inference time.
GWN is also noteworthy because it remained to be the state-of-the art according to the most recent benchmark study \cite{jiang2021dltraff}.
For this reason we chose GWN as the backbone forecasting model in this paper.

In our prior studies, G-SWaN \cite{prabowo2023GSWaN, PRABOWOArian2022StDL}, we observed that each sensor exhibits unique dynamics. Building upon this insight, we enhanced GWN by introducing a novel graph neural network called spatial graph transformer, which effectively captures these unique dynamics. Motivated by this, our current work directly addresses the question of how to learn these individual dynamics for new roads. We also further investigated the limitations of graph attention mechanism in traffic forecasting and explored message-passing mechanism instead \cite{prabowo2023MPNN4TrafficForecasting}.

MTGNN \cite{wu2020MTGNN} and GTS \cite{shang2021GTS} extended the idea of learning the latent topological connectivity from the data to any multivariate timeseries, instead of only traffic-related timeseries.
Conversely, because we are modelling traffic as multivariate timeseries in this paper, SCPT could easily be extended beyond traffic and spatiotemporal data, to any multivariate timeseries. An example of such use case is for electricity and gas retailers. It is important for them to be able to accurately forecast the usage of new clients with limited historical data and SCPT would be very well equipped to tackle this problem.

GWN, AGCRN \cite{DBLP:conf/nips/0001YL0020AGCRN}, GTS and SPGCL \cite{li2022SPGCL} used latent road embeddings that are learned during training.
Although these embeddings improved the model's forecasting accuracy, they prohibits inferences on new roads unseen in the training data because the latent embeddings for those roads have not been learned.
However, this paper shows that under the SCPT framework we can effectively infer the latent road embeddings on unseen roads during inference time using the spatial encoder.

ASTGCN and D\textsuperscript{2}STGNN \cite{shao2022d2stgnn} decoupled the traffic signals into multiple components and used specialized modules in their architecture to handle each signal component.
ASTGCN decoupled them to recent, daily-periodic, and weekly-periodic components.
Similarly, our SCPT framework decouples them to Markovian and periodic signals.
We call it Markovian, instead of recent, to highlight the property of the behaviors that our model is trying to capture.
Because the unseen roads setup is data scarce and the periodic signals is easier to model, we used discrete cosine transofrm (DCT) \cite{ahmed1974DCT} to model the period signals and to let the deep learning models to focus on the Markovian signals.

Noteworthy is also few-shot learning \cite{f1,zhang2020sopaccv,Lin_2018_ECCV,jeanie,zhang2022kernelized,hao_fsl,zhang2022time,Wang_2022_ACCV, udtw_eccv22, hao_fsl2} which can adapt to novel class concepts but it remains unclear how to apply few-shot learning on graphs.

\subsection{Contrastive learning}

There exist numerous approaches for the general-purpose graph contrastive learning \cite{wang2020understanding,zhu2021contrastive,zhu2021simple,zhang2022costa,sfa_yifei,zhu2021contrastive2}. However, to the best of our knowledge, there are only two works that implemented contrastive learning for traffic forecasting.
Both still used performance accuracy as their primary objective instead of other downstream tasks.
\cite{liu2022when} argued for the need of contrastive loss due to data scarcity in traffic forecasting.
Our unseen roads setup exacerbated this data scarcity problem and our results agree with them regarding the importance of contrastive learning.

SPGCL \cite{li2022SPGCL} used contrastive learning for neighbour connectivity selections, arguing that only roads that are similar in the contrastive embedding space should be topologically connected in the latent space.
This is like our integration of SCPT with our backbone where we use the output of the contrastively pre-trained spatial encoder to construct the latent topological connectivity.
The primary difference is that we allow this construction to be done on unseen roads during inference time.

FUNS-N \cite{roth2022FUNn} is the only prior work tackling unseen roads according to our knowledge although it was primarily tested on  simulated dataset.
In their paper, they called ‘unseen roads’ as ‘unobserved nodes’. 
Instead of motivating via newly constructed roads, they formulated their task as spatial imputation during spatio-temporal forecasting on a sensor network.
Another important difference with SCPT was that they chose a context-driven method (e.g., speed limits and road types) instead of a data-driven method.

\section{Methods: SCPT}

In this paper, we first introduce a novel setup where forecasting is performed on new roads unseen in the training data (Figure \ref{fig:vizabs}).
Here, the model (including the spatial encoder) is exposed to only a sample of sensor data in a traffic network.
Our main contribution is SCPT, a novel framework where a spatial encoder is pre-trained using contrastive loss, such that when it is attached to a forecasting model backbone, the latter generalizes well to new roads unseen in the training data.

This problem is formally defined in Subsection \ref{sec:def}.
The new setup requires a novel pre-training/training/validation/testing sets splitting strategy.
We call this the spatio-temporal split and it is described in Subsection \ref{sec:split}.
The pre-training of spatial encoder using contrastive loss is described in Subsection \ref{sec:ptc}.
The architecture of the spatial encoder itself, as implemented in this paper, is described in Subsection \ref{sec:encoder}.
With that being said, note that our framework is agnostic to the architecture of the spatial encoder.
The way the SCPT framework decouples traffic signal is described in Subsection \ref{sec:decouple}
Next, we introduce a new layer, called SGA, to effectively integrate the encoder output to existing traffic forecasting architectures.
This is described in Subsection \ref{sec:sga}.

Finally, to demonstrate the capabilities of the spatial encoder trained via the spatial contrastive pre-training framework, we combine it with GWN \cite{wu2019GraphWaveNet} as the forecasting model backbone.
The details of how to integrate SCPT frameworks to GWN and other backbones in general are described in Subsection \ref{sec:gwn}.

\subsection{Problem Definition: Forecasting on unseen roads}
\label{sec:def}

Figure \ref{fig:vizabs} shows how traffic signals are generated by sensors installed on roads in a traffic network.
This can be formulated as a multivariate timeseries where the traffic signals from the sensors forms a timeseries.
An alternative abstraction is for a sensor to be formulated as a node in a spatio-temporal graph.
The edges in this graph capture the spatial connections and relationships between sensors, reflecting either the physical proximity between sensors and road segments or the similarity of traffic signals in the latent space.
From these perspectives, we can use the terms node, sensor, and road interchangeably depending on the context.

The traffic dataset, denoted as
$\mathbf{X} \in \mathbb{R}^{M \times K}$,
is represented as a tensor.
Here, $M$ represents the number of traffic sensors in the dataset, and
$\mathcal{M}$ is the set of all road segments or sensors in the traffic network.
Thus, $M = \vert \mathcal{M} \vert $.
The dimension $K$ corresponds to the number of timesteps in the dataset.
For brevity, we assume there is only one traffic metric per timestep per sensor.
However, our method generalizes to multiple traffic metrics.
Each data point
$\mathbf{x}_{\mathcal{N}, k} = \mathbf{X}_{\mathcal{N},k:k+L}$
is a tensor, where 
$N = \vert \mathcal{N} \vert$ is the number of traffic sensors in the data point and where
$\mathcal{N} \in \mathcal{M}$
and, thus,
$N \leq M$,
and $k$ is the index of the timestep,
and
$L$ is the number of timesteps in the data point.
The tensor $\mathbf{X}$ is the traffic metric at a particular road, in a particular timestep.

As road networks are abstracted to spatio-temporal graphs with weighted edges, the topological connectivity between nodes is represented by a sparse adjacency matrix ($\mathbf{A}$).
The adjacency matrix is normalized between zero and one.
An edge with higher weight means that the nodes are closer together, while lower weight means that the nodes are further apart.
This topological formulation works both in the physical and latent spaces.

The classical traffic forecasting task is a multi-step forecasting problem formalized as follows:
$\mathbf{h}(\mathbf{x}_{\mathcal{M},k}, \mathbf{A})=\mathbf{x}_{\mathcal{M},k+L+F}$
where $F$ is the forecasting horizon.

However, since our model has not been trained on the data from all the roads in $\mathcal{M}$, we augment the forecast using a latent representation of the historical data produced by a spatial encoder $\mathbf{E}(\cdot)$.
Thus, our final formulation is as follows:
$\mathbf{h}(
\mathbf{x}_{\mathcal{M},k},
\mathbf{A},
\mathbf{E}(\mathbf{X}_{\mathcal{M},:k-1})
)=\mathbf{x}_{\mathcal{M},k+L+F}$

\subsection{Spatio-temporal split}
\label{sec:split}

\begin{figure}[htbp]
    \centering
    \includegraphics[width=\columnwidth]{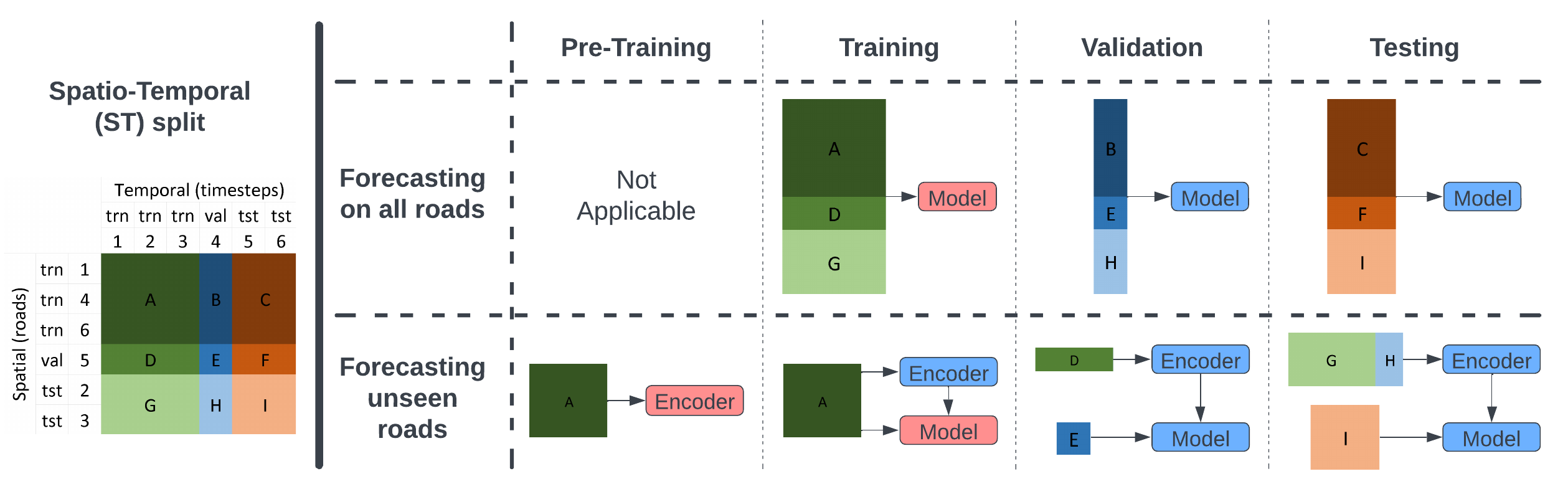}
    \caption{
    The ST splitting strategy divides the dataset into nine subsets (left side), while the right side illustrates the usage of different subset combinations at different stages.
    }
    \label{fig:split}
\end{figure}

When it comes to training/validation/testing splitting strategies, classical traffic forecasting where all roads are seen during training follows the setup in a typical timeseries.
The usual random shuffling is not appropriate for timeseries as training can only be done on historical data, and inferences are typically made on future data.
Temporally, the validation set must be at the future of training set, and the testing set is at the future of the validation set.
Spatially, the whole set contains all of the sensors.
This is shown in the top row of the right hand side of Figure \ref{fig:split}.

However, in our setup (bottom row) we are performing inference (testing) on roads that the model has never seen.
We call this new strategy a spatio-temporal (ST) split.
Here, we splits the dataset into nine subsets as shown at the left.
The top row shows the typical split a traffic forecasting setup where all roads are seen during training.
Meanwhile, the bottom row shows the new setup where not all roads are seen during training and forecasting is done on unseen nodes during inference (testing) time.
The encoder and forecasting models are exposed to different splits of the dataset during different stages.
When the encoder and model are colored red, they are being trained; but when they are blue, their weights are frozen for inference.
Note that the index of each timestep is in sequence, but the road index is random.


During pre-training, the encoder is trained on set $A$.
Then, it is validated on set $A \cup B \cup D \cup E$ (not shown in the figure).
This is to ensure that the encoder can generalize when it receives data from previously unseen sensors.
From this point onward, the encoder's weights are frozen.
The next subsection explains the pre-training procedures in detail.

During training, the model is trained on set $A$ with additional input from the encoder.
From this point onward, the model's weights are frozen.
During validation, the model is validated on set $E$, which is temporally separate from sets $A$ and $D$ to ensure that the model is not overfitted on past data, but generalizes into the future.
However, the inputs to the encoder is set $D$, which represent the past historical data.
During testing, the model is tested on set $I$, which only includes roads that neither the encoder, nor the model, has seen before.
The historical data $G \cup H$ serve as the inputs to the encoder.

Regarding the sizes of each set, for the temporal split, we follow the typical non-random split with the ratio of 7/1/2 \cite{jiang2021dltraff}.
For the spatial split, we roughly follow the same ratio.
In contrast to the non-random temporal split, the spatial split is random.
In this work we use a uniform distribution.
Further analysis in Subsection \ref{sec:split_result} shows that even uniform sampling produces acceptable results.

\subsection{Pre-training using contrastive loss}
\label{sec:ptc}

\begin{figure*}[htbp]
    \centering
    \includegraphics[width=\textwidth]{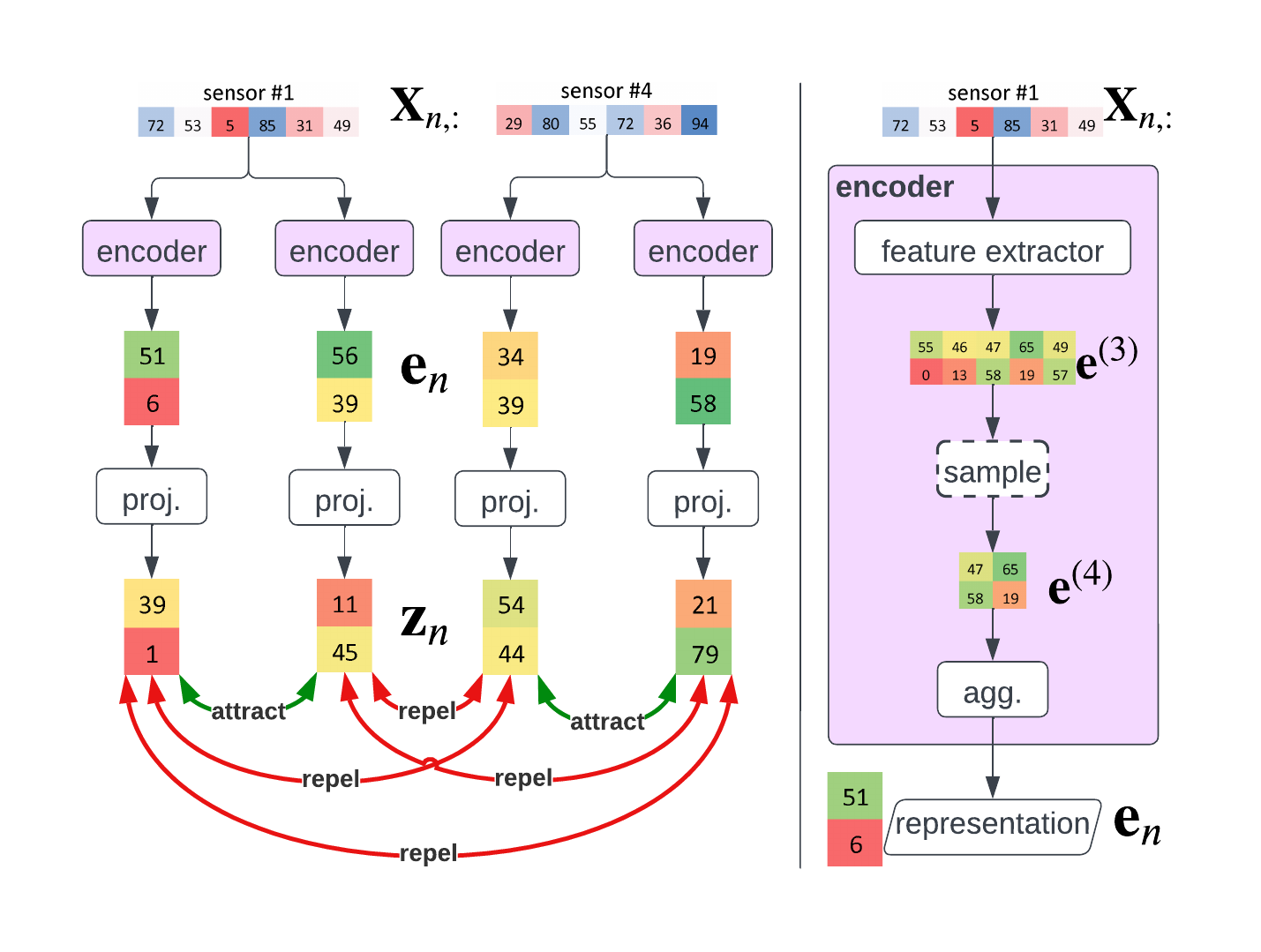}
    \caption{
    On the left, the use of contrastive loss to pre-train the spatial encoder is depicted, while on the right, the framework of the (spatial) encoder is illustrated.
    }
    \label{fig:scpt}
\end{figure*}

This section describes the pre-training procedure for the spatial encoder.
We use set $A$ (Figure \ref{fig:split}) as the pre-training data.
The pre-training using contrastive loss, as well as the architecture of the encoder, is shown in Figure \ref{fig:scpt}.
Adapted from SimCLR \cite{chen2020simclr} to increase training efficiency of traffic forecasting, it learns spatial representations by maximizing agreement of representations generated by a stochastic spatial encoder via a contrastive loss.

We randomly sample a minibatch of $N$ sensors from set $A$.
The spatial encoder takes each sensor's historical data twice and outputs two different representations of the sensor.
The representations are different because the spatial encoder is stochastic.
More formally:
$$
\mathbf{e_n}
\sim \mathbf{E}( \mathbf{X}_{n, :} )
;
\mathbf{e'_n}
\sim \mathbf{E}( \mathbf{X}_{n, :} )
$$
where 
$\mathbf{E}(\cdot)$ is the spatial encoder,
$\mathbf{X}_{n, :} \in \mathbb{R}^K $ is the historical data of sensor $n$,
$K$ is the number of timesteps in the set,
$\mathbf{e_n}, \mathbf{e'_n} \in \mathbb{R}^D$ are the two different representations of sensor $n$, 
and
$D$ is the embedding dimension.
The details of the spatial encoder will be described in the next subsection.

Although we want the two representations
$\mathbf{e_n}, \mathbf{e'_n}$
to be similar to each other,
following \cite{chen2020simclr}, we define the contrastive loss on the output of the projection head $p(\cdot)$.
We use a fully connected layer $FC(\cdot)$ as the projection head (proj. in Figure \ref{fig:scpt}):
$$
\mathbf{z}_n = p(\mathbf{e_n}) = FC(\mathbf{e_n})
$$

Finally, we use the normalized temperature-scaled cross entropy (NT-Xent) loss function \cite{chen2020simclr}, which is also known as InfoNCE \cite{oord2018CPC}.
The goal of this loss function is to ensure that the representations coming from the same sensor are similar to each other, while representations from different sensors are different.
Temperature is a hyperparamter of this loss function.
In section \ref{sec:temp} we show our framework is robust across a wide range of temperature.

\subsection{Spatial encoder}
\label{sec:encoder}

The spatial encoder $\mathbf{E}(\mathbf{X}_{n, :})$ extracts a latent representation of a road $\mathbf{e}_n$ from its historical data $\mathbf{X}_{n, :}$ in a stochastic manner.
Our spatial contrastive pre-training framework is agnostic to the choice of the spatial encoder architecture.
In this work, we pick simple architecture roughly based on the feature extractor for graph for time series (GTS) \cite{shang2021GTS}.
In the rest of this section, we describe our implementation of the spatial encoder as shown on the right in Figure \ref{fig:scpt}.

First, we pass the sensor's historical data $\mathbf{X}_{n, :}$ to a feature extractor.
The feature extractor consists of a sequence of three dilated convolutional layers \cite{vandenoord16_WaveNet}.
The windows are of sizes: 13, 12, and 24;
while the strides are: 1, 12, and 24, respectively.
Since each timestep in all of the datasets are 5 minutes long, these choices are made such that the output of each layer has the receptive field of sizes: 1 hour, 2 hours, and 1 day, respectively, and the final convolution output has no overlap between days.
The first convolution projects the data to a latent space of size $D$, which remains constant for the rest of the framework unless noted otherwise.
In between the convolutions there is a ReLU activation and also a batch normalization layer.
Formally:
\begin{align}
\mathbf{e}^{(1)} &
= BN^{(1)} \left( ReLU \left(
conv_{size=13, stride=1} \left(
\mathbf{X}_{n, :}
\right) \right) \right)
\\
\mathbf{e}^{(2)} &
= BN^{(2)} \left( ReLU \left(
conv_{size=12, stride=12} \left(
\mathbf{e}^{(1)}
\right) \right) \right)
\\
\mathbf{e}^{(3)} &
=
conv_{size=24, stride=24} \left(
\mathbf{e}^{(2)}
\right)
\end{align}
Note that the sizes of $\mathbf{e}^{(i)}$ depends on the size of set $A$ and they are also different for every layer due to the convolution dilation.

To extract multiple representations of sensor $n$, we sample from the latent space $\mathbf{e}^{(3)}$.
Additionally, this sampling process also reduces the dimension from the thousands of timesteps in the raw data to a smaller latent space.
Unlike the feature extractor in \cite{shang2021GTS} that uses a fully connected layer, this sampling process allows the encoder to extract representations from sensors with different lengths of historical data.
This offers a practical advantage when we are inferring on newly installed sensors on newly constructed roads with limited historical data.
Thus it is possible to perform inferences on new roads with only 2 days of data.
We uniformly sampled half of the days:
$$
\mathbf{e}^{(4)} = 
sample(\mathbf{e}^{(3)})
$$
such that
$\mathbf{e}^{(3)} \in \mathbb{R}^{D \times K^{(3)}}$,
$\mathbf{e}^{(4)} \in \mathbb{R}^{D \times K^{(4)}}$,
and
$K^{(4)} = K^{(3)}/2$
.

Finally, to aggregate the sampled representation, inspired from graph isomorphism network (GIN) \cite{xu2018GIN}, we take the $1^{st}$,
$2^{nd}$, and
$\infty^{th}$
statistical moments of
$\mathbf{e}^{(4)}$.
Then, we vectorize along the latent dimension and apply a fully connected layer to reduce the dimension back to $D$.
The output is a latent vector representation $\mathbf{e}_n$ of road $n$.
There are also ReLU activation and batch normalization layers as follows:
$$
\mathbf{e}_n
=
BN^{(4)} ( ReLU ( FC ( BN^{(3)} (
$$
$$
\mu(\mathbf{e}^{(4)})
\doubleplus
std(\mathbf{e}^{(4)})
\doubleplus
\texttt{max\_pool}(\mathbf{e}^{(4)})
$$
$$
) ) ) )
$$
where
$\mu(\cdot)$ is a mean pooling layer,
$std(\cdot)$ is a standard deviation pooling layer,
$\texttt{max\_pool}(\cdot)$ is a maximum pooling layer,
and
$\doubleplus$ is a concatenation operation along the latent dimension.

\subsection{SGA layer}
\label{sec:sga}

There are many different ways to integrate the output of the encoder into different layers in the forecasting architecture.
Here, for simplicity, we opt for summation.
However, a naive summation is problematic because it injects the same amount of spatial information all the time.
The amount of spatial information that a layer needs might differ between layers.
For this reason, we used gating mechanism with a scalar per sensor to decide how much spatial information should be added at each layer.

In brief, $SGA(\cdot)$ layer adds the latent representation vector
$\mathbf{e}_n$
to the activation $\mathbf{h}_n^{(l)}$ of the $l^{th}$ layer of the model, weighted by a coefficient
$c_n(\mathbf{h}_n^{(l)} , \mathbf{e}_n)$ that is unique for every sensor $n$.
More formally:
$$
\mathbf{h}_n^{(l+1)}= 
SGA(\mathbf{h}_n^{(l)} , \mathbf{e}_n)=
\mathbf{h}_n^{(l)} + c_n(\mathbf{h}_n^{(l)} , \mathbf{e}_n) \mathbf{e}_n
$$
where
$\mathbf{h}_n^{(l)} \in \mathbb{R}^D$ is the activation of the $l^{th}$ layer for sensor $n$ in the forecasting model,
$\mathbf{e}_n \in \mathbb{R}^D$ is the latent representation of sensor $n$ (the output of the frozen spatial encoder),
and
$c_n(\cdot) \in \mathbb{R}$ is the weight for sensor $n$ at layer $l$.
We calculate the weight $c_n(\cdot)$ using a multi-layer perceptron (MLP) with one hidden layer, ReLU activation, and wrap it under a sigmoid $\sigma(\cdot)$ to ensure that the weight is between 0 and 1:
$$
c_n(\mathbf{h}_n^{(l)} , \mathbf{e}_n) \in \mathbb{R} = 
\sigma \left( FC^{(2)} \left( ReLU \left( FC^{(1)} \left(
\mathbf{h}_n^{(l)} \doubleplus \mathbf{e}_n
\right) \right) \right) \right)
.
$$
The final fully connected layer $FC^{(2)}$ has an output size of $1$ to ensure that the output is a scalar.

\subsection{Traffic signal decoupling}
\label{sec:decouple}

Data scarcity has been identified as an essential issue in traffic forecasting \cite{liu2022when}.
The forecasting on unseen roads setup only exacerbated the data scarcity issue.
To address this, we decouple traffic signals to periodic Markovian signals.
The periodic signals are easy to model, i.e. traffic on Wednesday morning rush hours are similar to traffic from other Wednesday mornings, but is very different than traffic during Friday evening.
In contrast, the Markovian signals capture the complex spatio-temporal correlations between the recent past traffic from nearby roads.

Typically, deep learning models are able to learn the periodic signals effectively without compromising their capabilities to learn the Markovian signals at the same time.
However, due to the data scarcity issue in the new setup, we argue that it is better for the spatial encoders to only learn the Markovian signals.
We enforced this by decoupling the traffic signals.

Formally, we can decouple the traffic signals as follows:
$
\mathbf{x}_k = \mathbf{s}(k) + \hat{\mathbf{x}}_k
$
where
$\mathbf{s}(k)$
is the periodic signals since they are only dependant on the timestep index alone and
$\hat{\mathbf{x}}_k$
is the remaining Markovian signals.
Through this decoupling, the problem formulation becomes:
$\mathbf{h}(
\mathbf{x}_{k}, \hat{\mathbf{x}}_{k},
\mathbf{A},
\mathbf{E}(\hat{\mathbf{X}}_{:k-1})
)
+\mathbf{s}(k)
=\mathbf{x}_{k+L+F}$
as we ignore the spatial dimension in the notations for brevity.

We determined the periodic signals
$\mathbf{s}(k)$
by using road-wise DCT \cite{ahmed1974DCT} transform on the training set
$A \cup D \cup G$,
keeping the low-frequency coefficients by setting the high-frequency coefficients to zeroes, and then inverting the transform.
The number of low-frequency coefficients to be kept depended upon the minimization of MAE of the reconstruction against the validation set
$B \cup E \cup H$.

\subsection{Integration with GWN}
\label{sec:gwn}

\begin{figure}[htbp]
     \centering
         \includegraphics[width=0.8\textwidth]{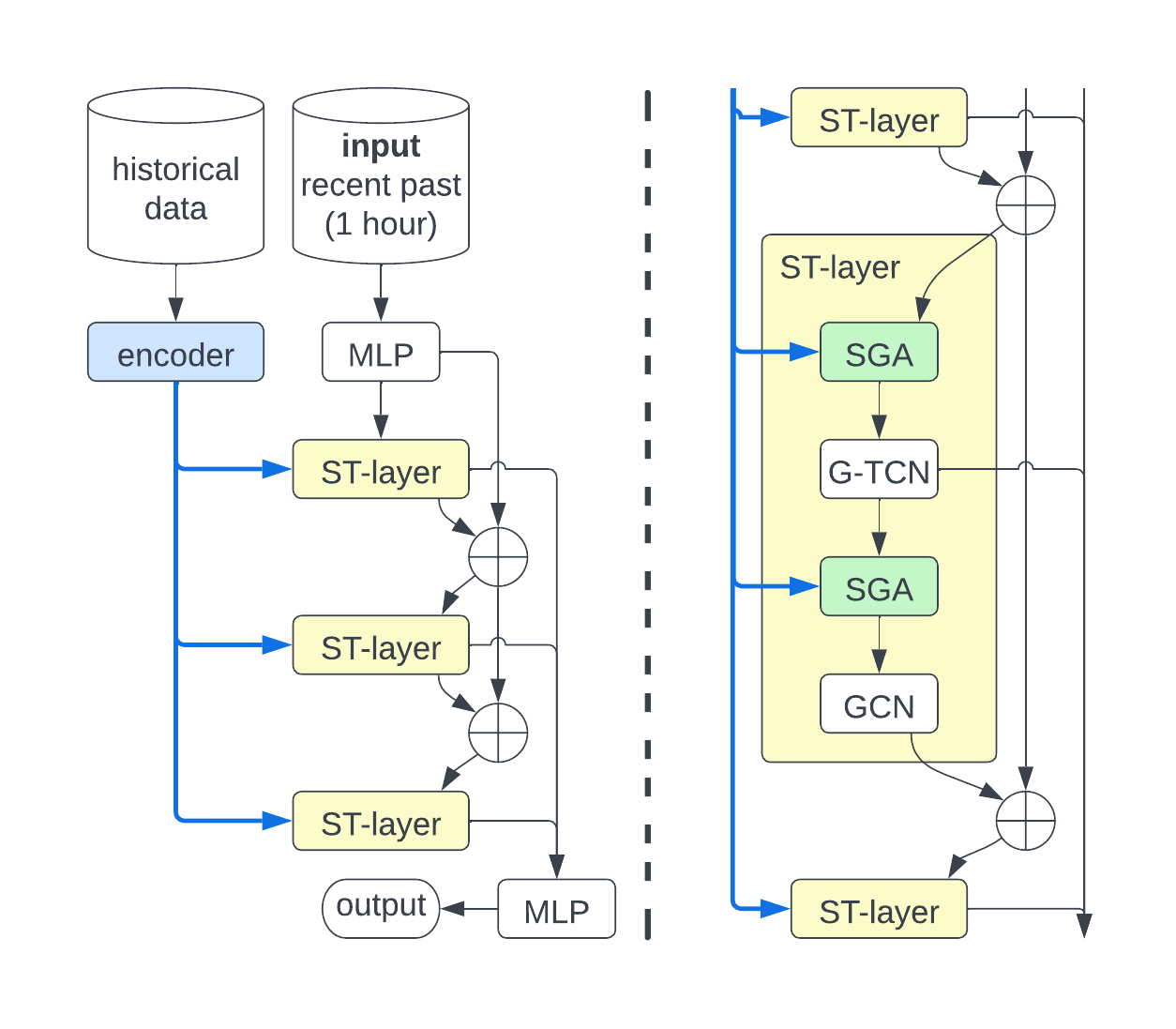}
    \caption{
    The left side of the figure illustrates the flow of outputs from the spatial encoder (blue) into the spatio-temporal (ST) layers (yellow). On the right side, the usage of Spatially Gated Addition (SGA) to integrate spatial information from the spatial encoder into the input of the G-TCN and GCN layers within the ST-layer is depicted.
    }
    \label{fig:gwn}
\end{figure}

To demonstrate the capabilities of the spatial contrastive pre-training framework, we combine it with GWN \cite{wu2019GraphWaveNet} as the forecasting model.
We picked GWN because it is the current state-of-the-art architecture according to the latest benchmark study \cite{jiang2021dltraff}.
The architecture of GWN, as well as the details of the integration with our framework, are illustrated in Figure \ref{fig:gwn}.
We are using the code implementation by the benchmark study \cite{jiang2021dltraff} that made them publicly available at \url{github.com/deepkashiwa20/DL-Traff-Graph}.

In brief, GWN consist of an initial MLP encoder, a sequence of ST-layers (in yellow), and a final MLP decoder.
Each ST layer consists of a G-TCN layer \cite{van2016wavenet} and GCN layer \cite{li2018dcrnn_traffic}.
There are also residual connections that go into the ST layers, and skip connections from the output of ST layers.
The skip connections are aggregated with concatenation before going into the final MLP decoder.
Integration with adaptive adjacency matrix is not shown.

The left hand side of Figure \ref{fig:gwn} shows that to integrate the outputs of the spatial encoder (in blue) trained using spatial contrastive pre-training to GWN, the outputs are given to ST-layers (in yellow) of GWN.
The right side shows that the components of an ST-layer are made up of gated temporal convolutional (G-TCN) layers and GCN layers.
We use SGA (in green) described in the previous subsection to inject the spatial information to the input of the G-TCN and the GCN layers.
Note that the weights of the spatial encoder are frozen during the training and inference of the GWN.

There are many other alternatives to our strategy of injecting spatial information into the spatial and temporal modules.
One naive approach is to treat it as a positional encoding at the start with the initial MLP.
However, we picked this strategy instead so that the models are not encumberred by forcing it to "remember" the entire spatial information that is only given once at the beginning.
Moreover, different layers might also require different "amount" of spatial information.
The SGA layer allows the network to modulate the "amount" of spatial information that gets injected at each layer.

Additionally, GWN used an adaptive adjacency matrix $\mathbf{A}_{adp}$ constructed via learned road embeddings
$\mathbf{A}_{adp} = 
SoftMax(ReLU(\mathbf{r}_1,\mathbf{r}_2^T))$ where
$\mathbf{r}_1$ and $\mathbf{r}_2$
are the road embeddings.
Because the road embeddings are learned during training, GWN does not allow the use of adaptive adjacency matrix in an unseen roads setup.
However, under the SCPT framework, we can use the outputs of the spatial encoder to infer the node embeddings via an MLP with one hidden layer
$\mathbf{r} =
FC(ReLU(FC(\mathbf{e})))$.
Each FC had their own unique set of weights and were not shared.

\section{Experiments}

\subsection{Dataset}

We utilized four real-world datasets for our experiments:
\textbf{METR-LA},
\textbf{METR-LA},
\textbf{METR-LA}, and
\textbf{METR-LA}.
The first three mentioned are popular datasets from the latest benchmark study \cite{jiang2021dltraff}.
The last one is the largest in the deep traffic forecasting literature \cite{mallick2020gpdcrnn}.
For more detailed description these datasets, including the detailed statistics, please refer to appendix \ref{sec:data_det}.

\subsection{Setups}

We use the same hyperparameters as the benchmark paper \cite{jiang2021dltraff}, including the use of Adam as the optimizer and Mean Absolute Error (MAE) as the loss function of the forecasting model.
The forecasting on unseen roads setup is more prone to overfitting on the trained roads.
Therefore, we use weight decay which value we treat as a hyperparameter.
Since we are comparing training time, we are training (and pre-training) for 100 epochs without the use of early stopping.
The latent spaces of the spatial encoder output has 32 dimensions, the same size as the GWN.
The middle layer of the MLP in SGA has 128 hidden nodes; the same amount as the hidden layer in the final MLP of GWN.
Similarly, the middle layers of the MLP in the construction of adaptive adjacency matrix also have 128 hidden nodes.
We used a standard scaler on the input of both the spatial encoder and GWN.
When the traffic signals are decoupled, the standard scaler is only applied after the decoupling.
All experiments were run on either NVIDIA Tesla P100 or V100 graphic cards.

\subsection{Results}

\begin{table}[]
\caption{
Performances evaluation of the SCPT framework using ST split.
In this setup, the models are trained on only 70\%, validated on 10\%, and tested on 20\% of the roads.
This table shows the average performance across 12 timesteps (1 hour) on the 20\% of the roads that are unseen during the training.
$\Delta$(\%) denotes the percentage of error reduction.}
\label{tab:result}
\centering
\begin{tabular}{@{}cl|lll@{}}
\toprule
Dataset                     & Methods      & RMSE    			& MAE    			& MAPE    			 \\ \midrule
\multirow{3}{*}{METR-LA}    & GWN          & 10.3405 $\pm$ 0.2634 & 4.7373 $\pm$ 0.1618 & 12.2677 $\pm$ 0.8058 \\
                            & GWN+SCPT	   & \textbf{10.0385} $\pm$ 0.2112 & \textbf{4.5645} $\pm$ 0.1556 & \textbf{11.5002} $\pm$ 0.8007 \\
                            & $\Delta$(\%) & 3\%     			& 4\%    			& 6\%     			 \\ \midrule
\multirow{3}{*}{PeMS-BAY}   & GWN          & 4.5059 $\pm$ 0.1613  & 2.0126 $\pm$ 0.1037 & 4.7779 $\pm$ 0.4303  \\
                            & GWN+SCPT	   & \textbf{3.9658} $\pm$ 0.1266  & \textbf{1.8163} $\pm$ 0.0875 & \textbf{4.1358} $\pm$ 0.2740  \\
                            & $\Delta$(\%) & 12\%    			& 10\%   			& 13\%    			 \\ \midrule
\multirow{3}{*}{PeMS-D7(m)} & GWN          & 6.4635 $\pm$ 0.3103  & 3.4327 $\pm$ 0.1974 & 8.6896 $\pm$ 0.7844  \\
                            & GWN+SCPT	   & \textbf{5.6893} $\pm$ 0.2552  & \textbf{3.0794} $\pm$ 0.1448 & \textbf{7.6770} $\pm$ 0.6678  \\
                            & $\Delta$(\%) & 12\%    			& 10\%   			& 12\%    			 \\ \bottomrule
\end{tabular}
\end{table}

To empirically evaluate the capabilities of SCPT, our proposed framework, on forecasting on new roads unseen in the training data, we use the ST split setup and tested it on three real-world data.
We replicated each experiment 10 times with different seeds for model weight initialization, sampling in the spatial encoder, and the selection of roads in the ST splits.
The results are displayed in Table \ref{tab:result}, showing the averages and the standard deviations across the 10 seeds.

The results shows that the SCPT framework consistently improve the backbone baselines across all three datasets and all three metrics.
The performance gain are more pronounced in the PeMS-BAY and PeMS-D7(m) datasets when compared to the METR-LA dataset.
This can be attributed to the fact that the METR-LA dataset has the widest speed distribution as indicated in the standard deviation in Table \ref{tab:data}.
More importantly, the SCPT framework also consistently reduce the standard deviations across all three datasets and all three metrics.
Thus, the SCPT framework, not only offers better performances, but also higher reliability.

\subsection{Performance across forecasting horizons}
\label{sec:fh}

\begin{figure}[htbp]
    \centering
    \includegraphics[width=0.9\textwidth]{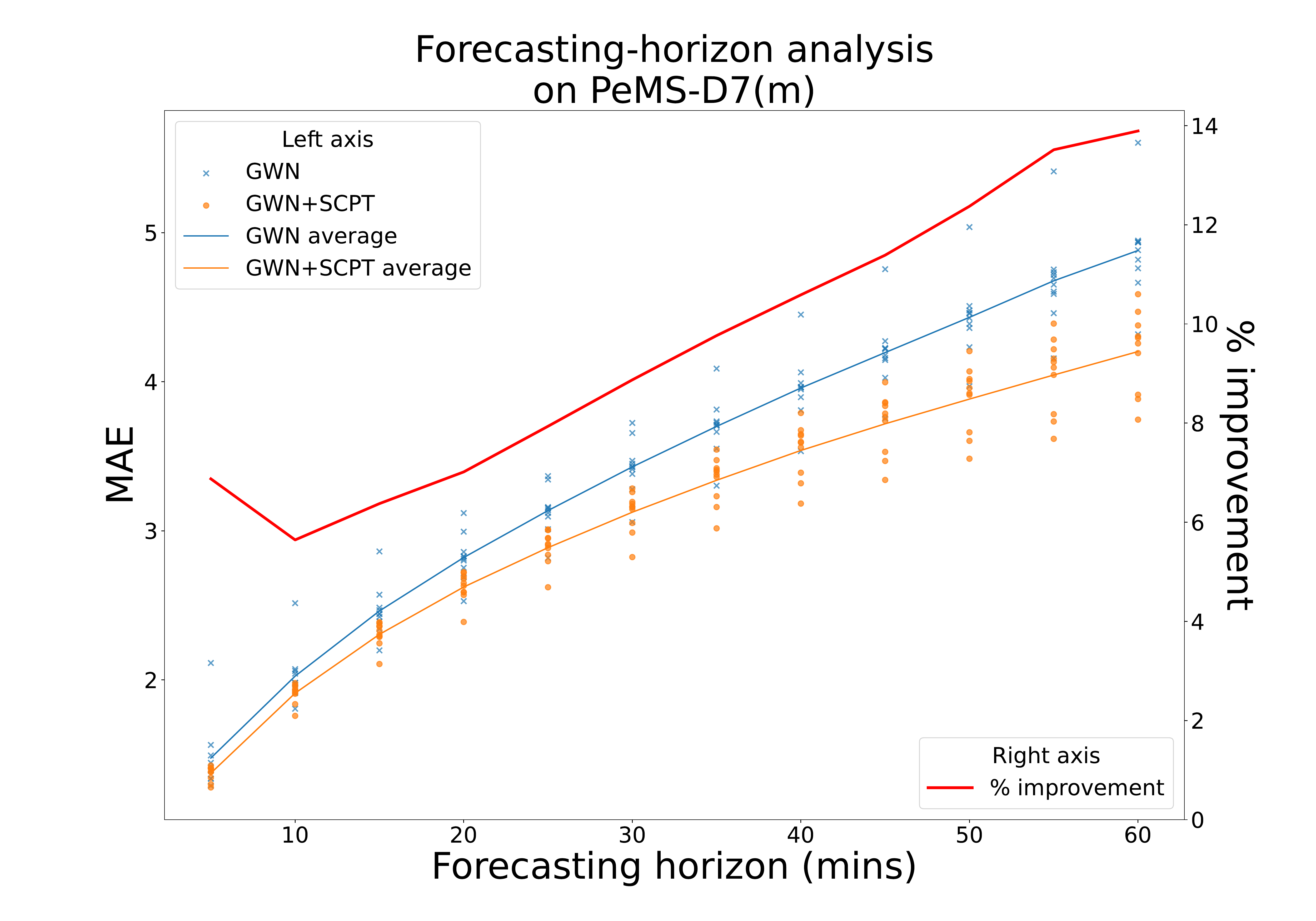}
    \caption{
    Performance across forecasting horizons.
    }
    \label{fig:fh}
\end{figure}

To further analyze the performance of the SCPT framework, we breakdown the results from the PeMS-D7(m) dataset in Table \ref{tab:result} based on forecasting horizons.
This is shown in Figure \ref{fig:fh}.
The average performance over the 10 seeds are show by the solid lines.
The thicker red line shows the percentage improvement of SCPT.
As expected, we see decreases in performance (increase in MAE) as the forecasting horizon increases i.e. it is harder to forecast further to the future.
The SCPT framework always improve the performance across all forecasting horizons.
More importantly, the performance gain brought by the SCPT framework also widens as forecasting horizon increases.
Although the average MAE improvement is 10\% as shown in \ref{tab:result}, at the furthest forecasting horizon (1 hour ahead), the improvement is nearly 14\%.
These results show the superiority of the SCPT framework.

\subsection{Ablation study}
\label{sec:ablat}

\begin{table}[]
\caption{
Ablation study on the PeMS-D7(m) dataset.
Every experiment is replicated five times (except the first and last ones).
The first row is the backbone baseline GWN and the last row is the full GWN+SCPT.
The \textbf{+} column shows the MAE reduction when compared to the GWN baseline (first row).
The \textbf{-} column shows the performance reduction when compared to the full SCPT framework.
}
\label{tab:ablt}
\centering
\begin{tabular}{@{}cccc|cccc@{}}
\toprule
\multicolumn{4}{c|}{Methods}     & \multicolumn{4}{c}{MAE}        \\ \midrule
SCPT & SGA & Decoupling & AdpAdj & mean       & std.       & \textbf{+}     & \textbf{-}     \\ \midrule
0    & 0   & 0          & 0      & 3.433	  & 0.197      & 0.000	& 0.353 \\
\checkmark    & 0   & 0          & 0      & 3.366      & 0.181      & 0.066	& 0.287 \\
\checkmark    &\checkmark  & 0          & 0      & 3.349      & 0.161      & 0.083	& 0.270 \\
\checkmark    & 0   &\checkmark         & 0      & 3.398      & 0.236      & 0.035	& 0.319 \\
\checkmark    & 0   & 0          &\checkmark     & 3.350      & 0.234      & 0.083	& 0.271 \\
\checkmark    & 0   &\checkmark         &\checkmark     & 3.249      & 0.255      & 0.184	& 0.169 \\
\checkmark    &\checkmark  & 0          &\checkmark     & 3.101      & 0.141      & 0.332	& 0.022 \\
\checkmark    &\checkmark  &\checkmark         & 0      & 3.406      & 0.187      & 0.026	& 0.327 \\
\checkmark    &\checkmark  &\checkmark         &\checkmark     & \textbf{3.079}	  & 0.145      & \textbf{0.353}	& \textbf{0.000} \\ \bottomrule

\end{tabular}
\end{table}

Here, we performed a complete ablation study to evaluate each component of the SCPT framework by iterating through all the possible combinations.
Table \ref{tab:ablt} shows the results of the ablation study.
Firstly, all of the entry of the \textbf{+} column is positive, meaning that all the individual components and combinations of them improve the baseline.

Based on this analysis, the most important component is AdpAdj, that is, the use of the output of the spatial encoders to construct the node embeddings in GWN.
GWN uses these node embeddings to construct the a adaptive adjacency.
Ablating AdpAdj effectively removed the adaptive adjacency from GWN as it has no mechanism to infer the node embeddings of new roads unseen in the training data.
The biggest lost of performance (\textbf{-} column) is found when ablating the AdpAdj, the second last row.

This finding agrees with the growing literature on the importance of inferring the latent topological connectivity of the traffic networks \cite{wu2020MTGNN, shang2021GTS, li2022SPGCL}.
This further highlights the capabilities of the SCPT frameworks to be easily integrated to various backbone models that require learned node embeddings and learning the latent topological connectivity of the traffic networks.

Additionally, our analysis revealed that the SGA component plays a critical role in integrating the output of the spatial encoder into the backbone network.
Notably, the addition of SGA alone exhibited comparable improvements in performance to the addition of AdpAdj alone.
Specifically, both configurations exhibited a gain of 0.083 in the third and fifth row of the \textbf{+} column.
This finding highlights the non-trivial nature of combining the output of the spatial encoder with the backbone network and emphasizes the importance of the SGA layer in effectively incorporating spatial information.

\subsection{Scalability to large dataset}
\label{sec:11k}

\begin{table}[]
\caption{Performance comparison on using the SCPT framework to train on a small sample (1\%) of roads to scale to a large dataset PeMS-11k(s).
$\Delta$(\%) denotes the percentage of error reduction.}
\label{tab:11k}
\centering
\begin{tabular}{@{}cccc|c@{}}
\toprule
Method:                            & GWN                 & GWN+SCPT            & $\Delta$(\%) & GP-DCRNN         \\ \midrule
\multicolumn{1}{c|}{RMSE}       & 5.6345 $\pm$ 0.7469 & \textbf{4.6741} $\pm$ 0.2089 & 17\%       &                  \\
\multicolumn{1}{c|}{MAE}        & 2.8241 $\pm$ 0.2840 & \textbf{2.4273} $\pm$ 0.2171 & 14\%       &                  \\
\multicolumn{1}{c|}{MAPE}       & 5.6345 $\pm$ 0.7469 & \textbf{4.6741} $\pm$ 0.2089 & 17\%       &                  \\
\multicolumn{1}{c|}{medianMAE12}   & 3.4554 $\pm$ 0.2343 & 3.2442 $\pm$ 0.3071 & 6\%        & \textbf{2.0200}           \\
\multicolumn{1}{c|}{Training time} & \textbf{00:16:39}            & 00:22:28            &            & 7 days, 22:34:53 \\ \midrule
\multicolumn{1}{c|}{\begin{tabular}[c]{@{}c@{}}Roads seen in\\ training (count)\end{tabular}} & \multicolumn{3}{c|}{111} & 11160 \\[3mm]
\multicolumn{1}{c|}{\begin{tabular}[c]{@{}c@{}}Roads seen in\\ training (\%)\end{tabular}}    & \multicolumn{3}{c|}{1\%} & 100\%
\end{tabular}
\end{table}

The capability to generalize forecasting performance well to unseen roads also opens up new avenues for more efficient traffic forecasting.
Training on large traffic network is costly. 
For example, training a model on one third of the entire California highway network took nearly 8 GPU-days \cite{mallick2020gpdcrnn}.
To make deep traffic forecasting more applicable for traffic management, a new scaling paradigm is required.
Instead of training a model on the entire traffic network, SCPT frameworks provide traffic managers with an attractive trade-off between forecasting accuracy and lower training cost by training only on a sample of roads while generalizing performance to the rest of the network.

To explore this new direction towards a more efficient traffic forecasting, we run a set of experiments where we train only on 1\% of the roads in PeMS-11k(s), but tested on the entire traffic network.
The results are shown in Table \ref{tab:11k}
We also use a metric called medianMAE12 to compare our results with \cite{mallick2020gpdcrnn}.
In medianMAE12, the MAE per road is calculated only for the 1 hour ahead forecasting horizon (12 timesteps).
Then, instead of averaging the MAE across all the roads, we take the median.

The results shows that training on a smaller sample of roads is a feasible way to increase the efficiency of traffic forecasting on larger traffic networks.
The GWN baseline achieved comparable performance to PeMS-BAY.
When combined with the SCPT framework, the performances increase even further.
The additional training cost is only 6 GPU-minutes.
The small degradation in performance is tolerable compared to more than hundreds fold savings in training time.

\subsection{Effect of randomness on sensors selection during the spatio-temporal split}
\label{sec:split_result}

\begin{figure}[htbp]
     \centering
     \begin{subfigure}[b]{.6\textwidth}
         \centering
         \includegraphics[width=\textwidth]{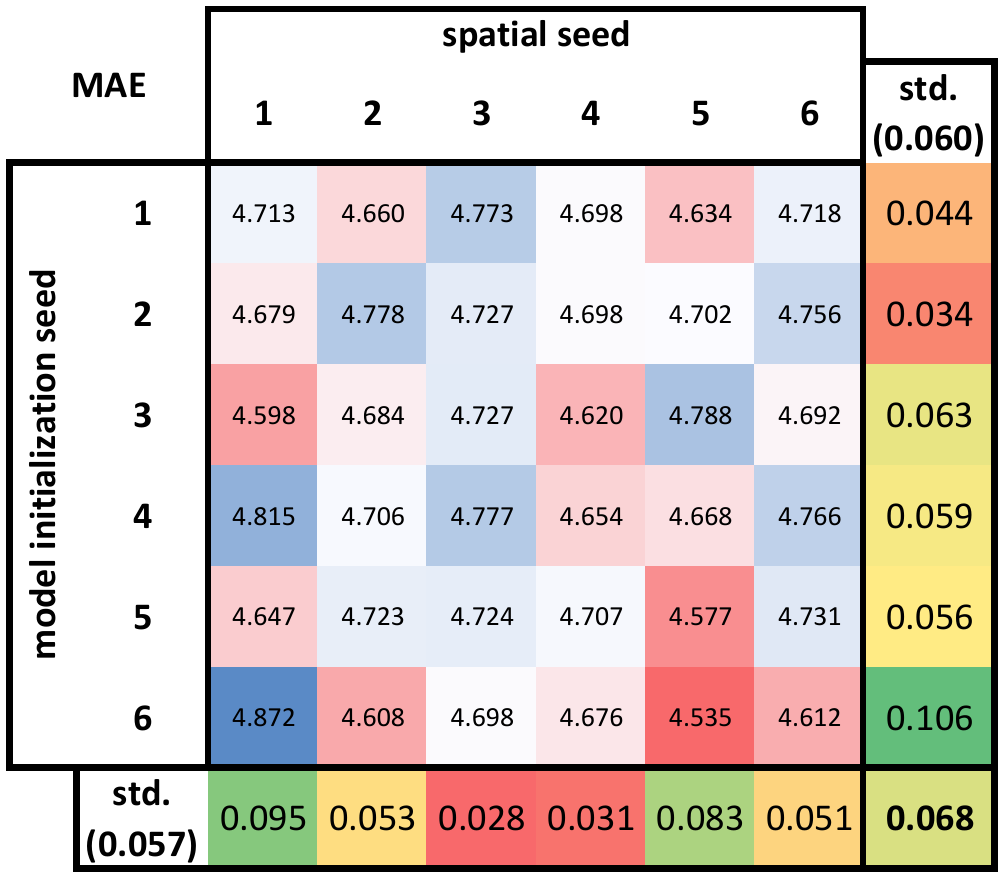}
     \end{subfigure}
    \caption{
    Analyzing the proposed framework's performance variance based on randomness in sensor selections in comparison with randomness in model weight initialization.
    }
    \label{fig:seed}
\end{figure}

In this set of experiments, we show that the impact of randomness in the ST split (Figure \ref{fig:split}) is greater than the effect of randomness in the models' weight initialization.
To this end, we separated the random seed used for weight initialization of the model and the random seed used for the ST split, and compared the impact.
We replicated this 36 times, each with a unique combination of model and splitting seeds.
The METR-LA dataset was used.

The result is shown in Figure \ref{fig:seed}.
Each value in a cell is the MAE of a run where the seed for the weight initialization is the heading of the row and the seed for sensor splitting is the heading of the column.
In every row, the model initialization seed is kept constant, so the spread captured by the standard deviation (on the right hand side of the figure) is solely due to randomness in the sensor selection process.
Similarly, for every column, the ST split is kept constant, so the spread is solely due to the randomness in the weight initialization process.
The average standard deviation due to randomness in the spatio-temporal splitting process (0.060) is similar to that of the model initialization process (0.057).
Notably, the combination of both sources of randomness (bottom right) does not significantly increase the spread of the performance (0.068).
This shows that the uniform random distribution over the sensors employed in the spatio-temporal split strategy does not affect the framework's performance significantly more than the randomness due to model's weight initialization.

\subsection{Computational cost of pre-training}

Spatial contrastive pre-training has a minimal impact on additional computational cost.
For the METR-LA dataset, when trained on all of the sensors, pre-training took less than 33 seconds, a negligible amount when compared to the training time of the forecasting model, which took 2 hours, 17 minutes.
Similarly, on PeMS-BAY, pre-training added an extra 58 seconds to the 5 hours and 27 minutes of the training process.

\section{Limitations and future works}


In the context of ST splits, our current approach of using uniform sampling has shown acceptable results (Section \ref{sec:split_result}).
However, we acknowledge the limitation of its naivety and the potential for further improvement.
In future work, we aim to explore more sophisticated sampling strategies, such as clustering-based approaches where we cluster the sensors based on the traffic signals, and sample only the centroids, or sampling based on the network topology.
These strategies have the potential to enhance the representation of the road network and improve the forecasting performance.

Our proposed SCPT framework is agnostic to the exact architecture of the spatial encoder.
The only requirement is that it behaves in a stochastic manner.
In this work, we proposed a simplistic WaveNet \cite{van2016wavenet} based encoder with statistical moments \cite{xu2018GIN} aggregations.
It is yet to be explored if performance can be improved by using more sophisticated domain generic encoders such as r-drop \cite{liang2021rdrop} and data2vec \cite{baevski2022data2vec}, or something more tailored to time series such as ts2vec \cite{yue2021ts2vec}.

In this paper, we employ different architectures for the spatial encoder and the backbone network.
Although using a single architecture for both components would be an elegant solution, it is not a straightforward task.
The spatial encoder's role is to summarize contrastive features from long time series (months to years), while the backbone network is designed to capture complex dynamics from shorter time series (minutes to hours).
Balancing these distinct requirements and developing a unified architecture would be an intriguing direction for future research.

We chose GWN \cite{wu2019GraphWaveNet} to be our backbone architecture because it is currently the best architecture available.
It is yet to be confirmed if the efficiency improvements brought by our framework can generalize to future state-of-the-art architectures.

Finally, we mentioned that forecasting using the SCPT framework can be generalized to any multivariate timeseries beyond just traffic.
More research are required in this direction.

\section{Conclusion}

In this paper, we proposed a novel task, that is, to perform traffic forecasting on new roads unseen in the training data.
To perform evaluations on this new task, we propose a novel setup called ST split.
Then we introduced the SCPT framework to train a spatial encoder on sensors' historical data.
Additionally, we implemented a simple spatial encoder to showcase our framework.
Next, we introduced an SGA layers, traffic signal decoupling, and a method to infer node embeddings using the output of the spatial encoder.
Finally, we evaluated our framework using GWN as the backbone forecasting model, on four real world datasets, showing consistent increases in performance and extensively analysed all of the components in the SCPT framework.

\section*{Conflict of interest statement}

The authors declare that this research was supported by the Data61/CSIRO PhD scholarship, RMIT Research Stipend Scholarship (RRSS), and the Australian Government RTP Scholarship.
We would like to also acknowledge the support of
the Investigative Analytics team (Data61/CSIRO)
and 
Cisco’s National Industry Innovation Network (NIIN) Research Chair Program.
The research utilized computing resources and services provided by
Gadi, supercomputer of the National Computational Infrastructure (NCI) supported by the Australian Government,
and
Bracewell, supercomputer of the Commonwealth Scientific and Industrial Research Organisation (CSIRO).

\bibliography{1bib}

\newpage
\appendix

\section{Datasets details}
\label{sec:data_det}

\begin{table}[htb]
\caption{Detailed statistics on the real world datasets.}
\label{tab:data}
\centering
\begin{tabular}{@{}ccrrrr@{}}
\toprule
\textbf{} &
  \multicolumn{1}{r}{\textbf{Dataset:}} &
  \multicolumn{1}{c}{\textbf{\begin{tabular}[c]{@{}c@{}}METR-\\ LA\end{tabular}}} &
  \multicolumn{1}{c}{\textbf{\begin{tabular}[c]{@{}c@{}}PeMS-\\ BAY\end{tabular}}} &
  \multicolumn{1}{c}{\textbf{\begin{tabular}[c]{@{}l@{}}PeMS-\\ D7(m)\end{tabular}}} &
  \multicolumn{1}{c}{\textbf{\begin{tabular}[c]{@{}l@{}}PeMS-\\ 11k(s)\end{tabular}}} \\ \midrule
\multirow{2}{*}{\rotatebox[origin=c]{90}{Spatial}}     & \multicolumn{1}{c|}{Nodes}                & 207       & 325        & 228       & 11,160      \\[1mm]
                             & \multicolumn{1}{c|}{Edges}                & 1,515     & 2,694      & 7,304     & 234,966     \\[1mm] \midrule
\multirow{5}{*}{\rotatebox[origin=c]{90}{Temporal}}    & \multicolumn{1}{c|}{Duration (timesteps)} & 34,272    & 52,116     & 12,672    & 25,632      \\
                             & \multicolumn{1}{c|}{Duration (days)}      & 121       & 150        & 61        & 89          \\
                             & \multicolumn{1}{c|}{Time start}           & 01-Mar-12 & 01-Jan-17  & 01-May-12 & 01-Feb-18   \\
                             & \multicolumn{1}{c|}{Time end}             & 30-Jun-12 & 31-May-17  & 30-Jun-12 & 30-Apr-18   \\
                             & \multicolumn{1}{c|}{Granularity (mins)}   & 5         & 5          & 5         & 5           \\ \midrule
\multirow{8}{*}{\rotatebox[origin=c]{90}{Speed (mph)}} & \multicolumn{1}{c|}{Min}                  & 0.00      & 0.00       & 3.00      & 3.00        \\
                             & \multicolumn{1}{c|}{Q1}                   & 57.13     & 62.10      & 57.50     & 62.60       \\
                             & \multicolumn{1}{c|}{Median}               & 63.22     & 65.30      & 64.10     & 65.10       \\
                             & \multicolumn{1}{c|}{Mean}                 & 58.46     & 62.62      & 58.89     & 63.14       \\
                             & \multicolumn{1}{c|}{Q3}                   & 66.50     & 67.50      & 66.70     & 67.80       \\
                             & \multicolumn{1}{c|}{Max}                  & 70.00     & 85.10      & 82.60     & 99.30       \\
                             & \multicolumn{1}{c|}{Standard Deviation}   & 20.26     & 9.59       & 13.48     & 9.01        \\
                             & \multicolumn{1}{c|}{Missing values}       & 8.82\%    & 0.00\%     & 0.00\%    & 0.00\%      \\ \midrule
\multirow{2}{*}{\rotatebox[origin=c]{90}{Size}}        & \multicolumn{1}{c|}{Entry}                & 7,094,304 & 16,937,700 & 2,889,216 & 286,053,120 \\
                             & \multicolumn{1}{c|}{Compressed (MB)}      & 54        & 130        & 6         & 2,235       \\ \bottomrule
\end{tabular}
\end{table}

We utilized four real-world datasets for our experiments. The \textbf{METR-LA} dataset was collected from loop detectors in Los Angeles, United States county highways. The \textbf{PeMS-BAY} dataset was collected by the California Transportation Agencies (CalTrans) Performance Measurement System (PeMS) using loop detectors located around the San Francisco Bay Area. Similarly, the \textbf{PeMS-D7(m)} dataset was also collected by CalTrans PeMS using loop detectors, but only during weekdays. The \textbf{PeMS-11k} dataset, which is the largest in the deep traffic forecasting literature, was also collected by CalTrans PeMS using loop detectors. However, to focus on spatial generalization and optimize resource utilization, we only utilized a two-month period from the original one-year-long dataset, referred to as \textbf{PeMS-11k(s)}.

The first three datasets mentioned are from the latest benchmark study \cite{jiang2021dltraff}, and they have been made publicly available on their GitHub repository \burl{github.com/deepkashiwa20/DL-Traff-Graph}. The last dataset is from \cite{mallick2020gpdcrnn} and is also available on their GitHub page \burl{https://github.com/tanwimallick/graph_partition_based_DCRNN}. For detailed statistics about these datasets, please refer to Table \ref{tab:data}.

\section{Temperature sensitivity}
\label{sec:temp}

One of the main hyperparameters introduced in our framework is the NT-Xent temperature.
We conducted an experiment to show that our framework is robust across a large range of temperature values.
This experiment was run using the METR-LA dataset.
Since temperature is a hyperparameter, we are analyzing the framework's performance on the validation set.

\begin{figure}[htbp]
\centering
\includegraphics[width=\textwidth]{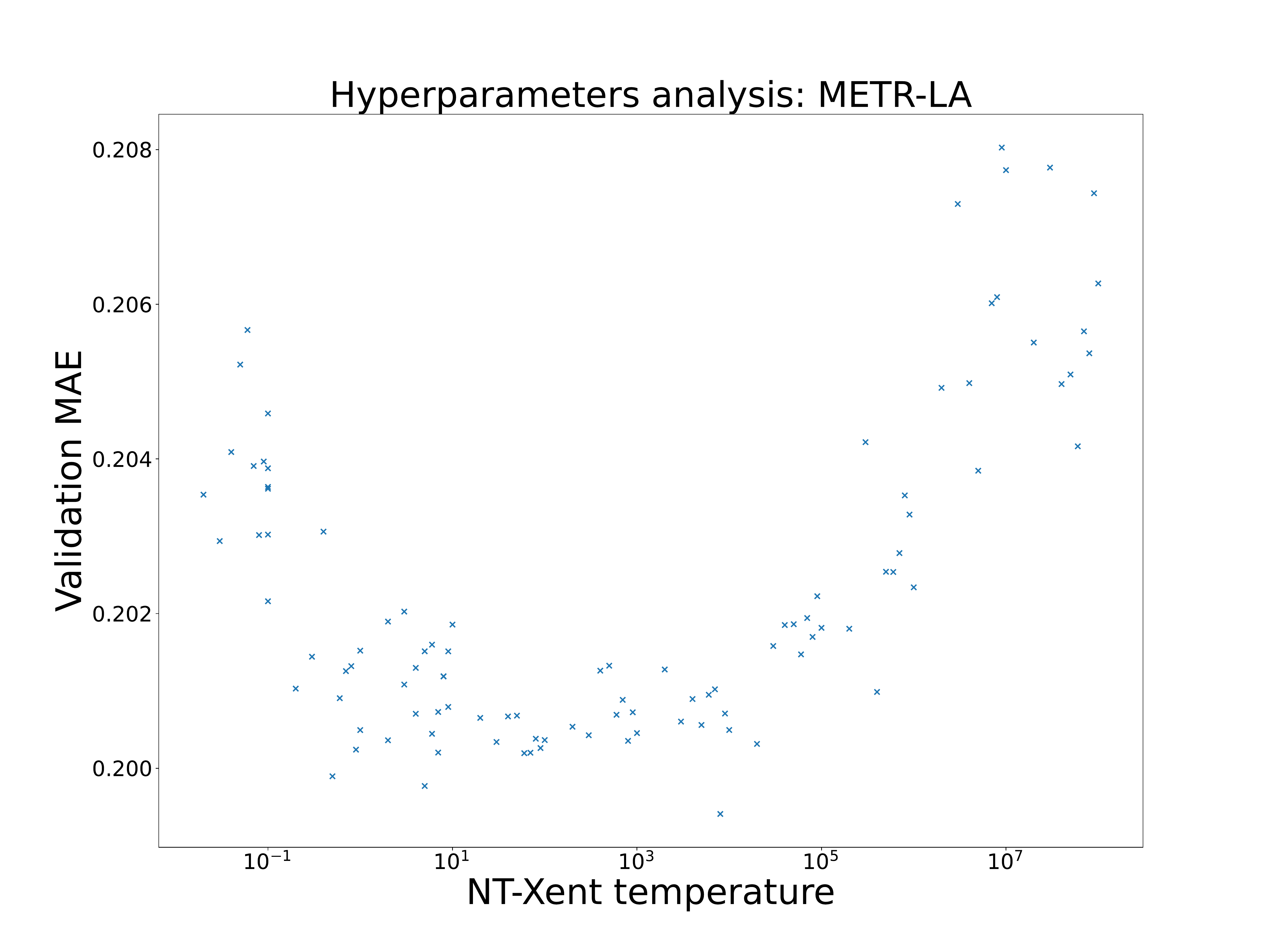}
\caption{Analyzing the framework's sensitivity against NT-Xent temperature hyperparameter.}\label{fig:temp1}
\end{figure}

\begin{figure}[htb]
\centering
\includegraphics[width=.8\textwidth]{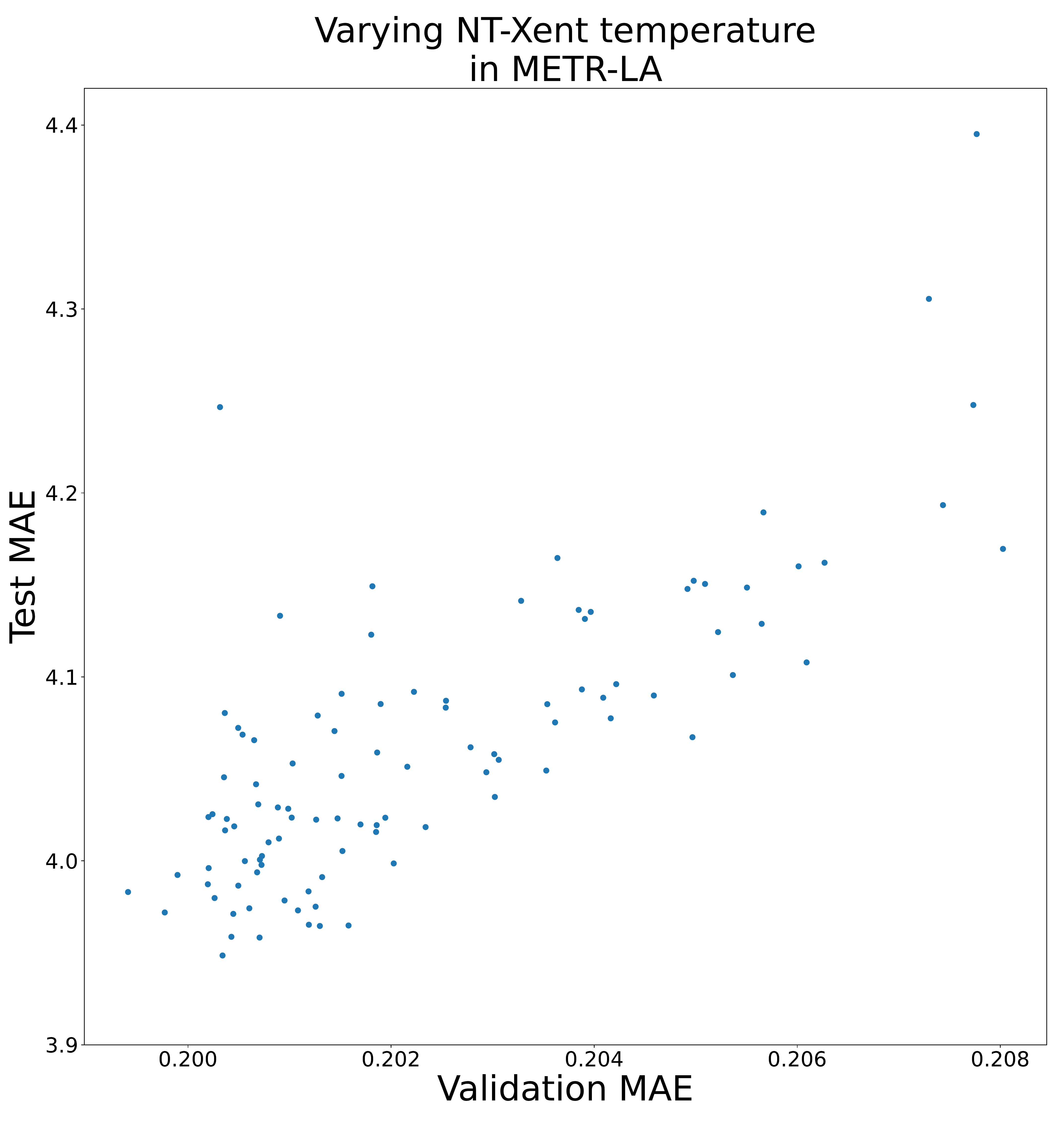}
\caption{Correlation validation and test MAE when verying NT-Xent temperature.}\label{fig:temp_val_tst}
\end{figure}

Figure \ref{fig:temp1} shows the result of this set of experiments.
It shows that the validation performance is stable across a wide range of temperature, from around $10^1$ to approximately $10^4$, which is about three orders of magnitude.
Additionally, the validation MAE is correlated with the test MAE over changes in temperature, as shown in Figure \ref{fig:temp_val_tst}, which shows that validation and test MAE are correlated against variation in NT-Xent temperature.
These results suggest that the robustness of the framework across different temperatures during hyperparameter optimization can reasonably transfer to the test performance.

\end{document}